\def\year{2020}\relax
\documentclass[letterpaper]{article} 
\usepackage{aaai20}  
\usepackage{times}  
\usepackage{helvet} 
\usepackage{courier}  
\usepackage[hyphens]{url}  
\usepackage{graphicx} 
\usepackage{graphicx}  
\usepackage{amsmath}
\usepackage{amssymb}
\usepackage{siunitx}
\usepackage{xspace}
\usepackage{filecontents}

\usepackage{tikz}
\usetikzlibrary{calc,arrows}
\usepackage{pgfplots}
\usepackage{pgfplotstable}
\pgfplotsset{compat=1.12}

\usepackage{csvsimple}

\newcommand{\ourTool}{GRT\xspace}
\newcommand{\ourToolCVC}{GRT+CVC4\xspace}

\newcommand{\grammar}{\ensuremath{G}\xspace}
\newcommand{\grammarSynth}{\ensuremath{\grammar^{\star}}\xspace}
\newcommand{\grammarCrit}{\ensuremath{\grammar_{crit}}\xspace}
\newcommand{\program}{P}
\newcommand{\constraints}{\ensuremath{C}\xspace}
\newcommand{\terminal}{\ensuremath{g}\xspace}
\newcommand{\synthesizesInTime}[1]{\rightsquigarrow_{#1}}
\newcommand{\synthesizes}{\rightsquigarrow}
\newcommand{\synthTime}[2]{\ensuremath{T^{#1}_{#2}}}
\newcommand{\satisfies}{\vdash}
\newcommand{\drop}[2]{\ensuremath{#1 \setminus #2}}
\newcommand{\setOf}[1]{\ensuremath{\pi(#1)}}

\newcommand{\trainingCrit}{\ensuremath{\mathcal{TR}_{crit}}}
\newcommand{\trainingTime}{\ensuremath{\mathcal{TR}_{time}}}

\newcommand{\reals}{\ensuremath{\mathbb{R}}}

\title{Grammar Filtering For Syntax-Guided Synthesis}
\author{
Kairo Morton,\textsuperscript{\rm 1}
William Hallahan,\textsuperscript{\rm 2}
Elven Shum,\textsuperscript{\rm 3}
Ruzica Piskac,\textsuperscript{\rm 2}
Mark Santolucito,\textsuperscript{\rm 2}\\
\textsuperscript{\rm 1}George School,
\textsuperscript{\rm 2}Yale University,
\textsuperscript{\rm 3}Deerfield Academy\\
mortonk@georgeschool.org, william.hallahan@yale.edu, eshum20@deerfield.edu, \\ ruzica.piskac@yale.edu, mark.santolucito@yale.edu
}

\begin{document}

\maketitle

\begin{abstract}
Programming-by-example (PBE) is a synthesis paradigm that allows users to generate functions by simply providing input-output examples.
While a promising interaction paradigm, synthesis is still too slow for realtime interaction and more widespread adoption.
Existing approaches to PBE synthesis have used automated reasoning tools, such as SMT solvers, as well as works applying machine learning techniques.
At its core, the automated reasoning approach relies on highly domain specific knowledge of programming languages.
On the other hand, the machine learning approaches utilize the fact that when working with program code, it is possible to generate arbitrarily large training datasets.
In this work, we propose a system for using machine learning in tandem with automated reasoning techniques to solve Syntax Guided Synthesis (SyGuS) style PBE problems.
By preprocessing SyGuS PBE problems with a neural network, we can use a data driven approach to reduce the size of the search space, then allow automated reasoning-based solvers to more quickly find a solution analytically.
Our system is able to run atop existing SyGuS PBE synthesis tools, decreasing the runtime of the winner of the 2019 SyGuS Competition for the PBE Strings track by 47.65\% to outperform all of the competing tools.
\end{abstract}

\section{Introduction}

The term ``program synthesis'' 
refers to automatically generating code to satisfy some specification. That specification describes {\emph{what}} the code should do, without going into details about {\emph{how}} it should be done. The specification could be given as a set of constraints~\cite{MannaW79,KuncakMPS10}, it can be deduced from the program and its environment~\cite{GveroKKP13,Feng:2017:CSC:3009837.3009851}, or it can be inferred from a large corpus~\cite{BalogGBNT17,SantolucitoZDSP17}.

One paradigm of program synthesis is called \textit{programming by example}~\cite{Cypher:1993:WIP:168080} (PBE). In the PBE approach, a user only provides a set of pairs of input-output examples that illustrate the desired behavior of the code. From these examples, the PBE engine should then generate code that generalizes from the examples to create a program which covers the unspecified examples as well.

The idea of automated code synthesis is an area of research with a long history (cf. the Church synthesis problem~\cite{Church1963-CHUAOR}). 
However, due to the problem's undecidability and high computational complexity for decidable fragments, for almost 50 years the research in program synthesis was mainly focused on addressing theoretical questions and the size of synthesized programs was relatively small. However, the state of affairs has drastically
changed in the last decade. By leveraging advances in automated reasoning and formal methods, there has been a renewed interest in software synthesis. The research in program synthesis has recently focused on
developing efficient algorithms and tools, and synthesis has even been used in industrial software~\cite{gulwani2011automating}. Today, machine learning plays a vital role in modern software synthesis and there are numerous tools and startups that rely on machine learning and big data to automatically generate code~\cite{codataWeb,BalogGBNT17}.

With numerous synthesis tools and formats being developed, it was difficult to empirically evaluate and compare existing synthesis tools. The Syntax Guided Synthesis (SyGuS) format language~\cite{alur2013syntax,sygusWeb} 
was introduced in an effort to standardize the specification format of program synthesis, including PBE synthesis problems. The SyGuS language specifies synthesis problems through two components - a set of constraints (eg input-output examples), and a grammar (a set of functions).
The goal of a SyGuS synthesis problem is to construct a program from functions within the given grammar that satisfies the given constraints. With this standardized synthesis format and an ever expanding set of benchmarks, there is now a yearly competition of synthesis tools~\cite{sygusCompetition2019}, which pushes the frontier of scalable synthesis further.

The SyGuS Competition splits synthesis  problems into tracks, for example PBE Strings or PBE BitVectors, assigning a different grammar for each track - and sometimes even varying the grammar within a single track.
As the grammar defines the search space in SyGuS, this allows benchmark designers to ensure problems are relatively in-scope of current tools.
However, when synthesis is deployed in real-world applications, we must allow for larger grammars that account for the wide range of use-cases users require~\cite{chi19}.
While larger grammars allow for more expressive power in the synthesis engine, it also slows down the whole synthesis process.

In our own experimentation, we found that by manually removing some parts of the grammar from the SyGuS Competition benchmarks, we can significantly improve synthesis times.
Accordingly, we sought to automate this process.
Removing parts of a grammar is potentially dangerous though, as we may remove the possibility of finding a solution altogether.
In fact, understanding the grammar's impact on synthesis algorithms is a complex problem, connected to the concept of overfitting~\cite{cav19overfitting}.

In this paper, we utilize machine learning to automate an analysis of a SyGuS grammar and a set of synthesis constraints.
We generate a large number of SyGuS problems, and use this data to train a neural network.
Given a new SyGuS problem, the neural network predicts how likely it is for a given grammar element to be critical to synthesizing a solution to that problem.
Our key insight is that, in addition to criticality, we predict how much time we expect to save by removing this grammar element.
We combine these predictions to efficiently filter grammars to fit a specific synthesis problem, in order to speed up synthesis times.
Even with these reduced grammars, we are still able to find solutions to the problems.

We implemented our approach in a modular tool, \ourTool, that can be attached to any existing SyGuS synthesis engine as a blackbox.
We evaluated \ourTool by running it on the SyGuS Competition Benchmarks from 2019 in the PBE Strings track.
We found \ourTool outperformed CVC4, the winner of the SyGuS Competition from 2019, reducing the overall synthesis time by $47.65\%$.
Additionally, \ourTool was able to solve a benchmark for which CVC4 timed out.

In summary, the core contributions of our work are as follows:
\begin{enumerate}
\item A methodology to generate models that can reduce time needed to synthesize PBE SyGuS problems.
In particular, our technique reduced the grammar by identifying which functions to try to eliminate to increase the efficiency of a SyGuS solver.
It also learns a model to predict which functions are critical for a particular PBE problem. 
\item A demonstration of the effectiveness of our methodology.  We show experiments on existing SyGuS PBE Strings track that demonstrates the speed up resulting from using our filtering as a preprocessor for an existing SyGuS solver.  Over the set of benchmarks, our techniques decreases the total time taken by synthesis by $47.65\%$.
\end{enumerate}

\section{Related}

One approach to SyGuS is to directly train a neural network to satisfy the input/output examples~\cite{andrychowicz2016learning,devlin2017robustfill,graves2014neural,joulin2015inferring,kaiser2015neural,ChenLS17a}.
However, such approaches struggle to generalize, especially when the number of examples is small~\cite{devlin2017neural}.
Some existing work~\cite{wang2018execution,bunel2018leveraging} aims to represent aspects of the syntax and semantics of a language
in a neural network.
In contrast to these existing approaches, which aim to outright solve SyGuS problems, our work acts as a preprocessor for a separate SyGuS solver.
However, one could also explore using our work as a preprocessor for one of these existing neural network directed synthesis approaches.
Other works have explored combining logic-directed and machine learning guided synthesis approaches~\cite{nye2019learning}.
This work sought to split synthesis tasks between generating high level sketches with neural networks, and fill in the holes of the sketch with an enumerative solver. 
Our work could be complementary to this, by assisting in pruning of the search space needed to fill in the holes.
 
Like our work, DeepCoder~\cite{BalogGBNT17} and Neural-Guided Deductive Search (NGDS)~\cite{kalyan2018neural}
identify pieces of a grammar that should be removed from the grammar.
However, in our parlance, these works only consider \textit{criticality}, which measures how important a part of the grammar is to completing synthesis.
Unlike our work, they do not consider the time savings from removing or keeping a part of the grammar.
NGDS~\cite{kalyan2018neural} does note that different models could be trained for different pieces of a grammar,
however, it provides no means of automating this process.
Rather, the user would have to manually elect to train individual neural networks for different grammatical elements.
Work by Si et al~\cite{si2018learning} aims to learn an efficient solver for a SyGuS from scratch, rather than, as in our work, acting as a preprocessor for a separate solver.
\section{Background}

A SyGuS synthesis problem is a tuple $(\constraints, \grammar)$ of constraints, \constraints, and a context-free grammar, \grammar.
In our case we restrict the set of constraints to the domain of PBE, so that all constraints are in the form of pairs $(i,o)$ of input-output examples.
We write \drop{\grammar}{\terminal} to denote the grammar \grammar, but without the terminal symbol \terminal.
The set of terminal symbols are the component functions that can be used in constructing a program (e.g. +, -, str.length).
We also use the notation, \setOf{\grammar}, to denote the projection of \grammar into its set representation, which is the set of the terminal symbols in the grammar.

The problem statement of syntax-guided synthesis (SyGuS) is; given a grammar, \grammar, and a set of constraints \constraints, find a program, $\program \in \grammar$, such that the program satisfies all the constraints -- $\forall c \in \constraints. \program \satisfies c$.
For brevity, we equivalently write $\program \satisfies \constraints$.
If our synthesis engine is able to find such a program in $t$ seconds or less, we write that $(\grammar, \constraints) \synthesizesInTime{t} \program$.
We use the notation $\synthTime{\constraints}{\grammar}$ to indicate the time to run $(\grammar, \constraints) \synthesizesInTime{t} \program$.
If the SyGuS solver is not able to find a solution within the timeout ($\synthTime{\constraints}{\grammar} > t$), we denote this as $(\grammar, \constraints) \not \synthesizesInTime{t} \program$.
We typically set a timeout on all synthesis problems of 3600 seconds, the same value of the timeout used in the SyGuS competition.
We write $(\grammar, \constraints) \synthesizes \program$ and $(\grammar, \constraints) \not \synthesizes \program$
as shorthand for $(\grammar, \constraints) \synthesizesInTime{3600} \program$ and $(\grammar, \constraints) \not \synthesizesInTime{3600} \program$, respectively.

We define \grammar as the grammar constructed from the maximal set of terminal symbols we consider for synthesis.
We call a terminal, \terminal, within a grammar, \textit{critical} for a set of constraints, \constraints, if $(\drop{\grammar}{g}, \constraints) \not \synthesizes \program$.
For any given set of constraints, if a solution exists with \grammar, there is also a grammar, \grammarCrit, that contains exactly the critical terminal symbols required to find a solution.
More formally, \grammarCrit is constructed such that
\begin{equation*}
(\grammarCrit, \constraints) \synthesizes \program \land
\forall g \in \grammarCrit. \ (\drop{\grammar}{g}, \constraints) \not \synthesizes \program 
\end{equation*}
Note that \grammarCrit is not unique.

The goal of our work is to find a grammar, \grammarSynth, where $\setOf{\grammarCrit} \subseteq \setOf{\grammarSynth} \subseteq \setOf{\grammar}$.
This will yield a grammar that removes some noncritical terminal symbols so that the search space is smaller, but still sufficient to construct a correct program.

\section{Overview}

\begin{figure}
\centering
\tikzstyle{module} = [style={draw,rectangle}, text width=2.2cm,align=center]
\tikzstyle{data} =   [style={draw,rectangle}, text width=2.2cm,align=center, fill=blue!5!white]
\usetikzlibrary{calc}

\begin{tikzpicture}[node distance=1.5cm,auto,>=latex']
    \node [data] (grammar) {Grammar \grammar};
    \node [data, right of=grammar, node distance=3cm] (constraints) {Constraints \constraints};
    \node [module, below of=grammar] (time) {Predict Time};
    \node [module, below of=constraints] (crit) {Criticality};
    \node [module, below of=crit] (combo) at ($(time)!0.5!(crit)$) {Combo};
    \node [data, below of=combo] (synthG) {Predicted Grammar \grammarSynth};

    \path[->] (grammar) edge (time);
    \path[->] (constraints) edge (time);
    \path[->] (grammar) edge (crit);
    \path[->] (constraints) edge (crit);

    \path[->] (time) edge (combo);
    \path[->] (crit) edge (combo);

    \path[->] (combo) edge (synthG);

\end{tikzpicture}
\caption{\ourTool uses the grammar \grammar and constraints \constraints to predict how critical each function is,
and the amount of time that would be saved by eliminating it from the grammar.
Then, it outputs a new grammar \grammarSynth, which it expects will speed up synthesis over the original grammar (that is, it expects that $T^C_{\grammarSynth} < T^C_\grammar$).}
\label{fig:model}
\end{figure}
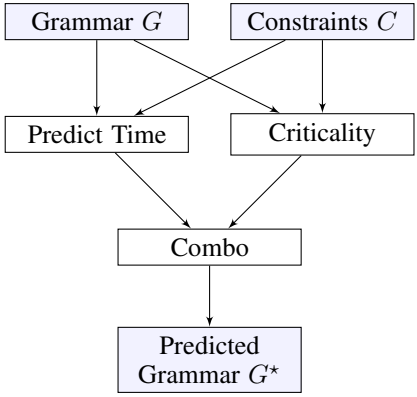

Our system, \ourTool, works as a preprocessing step for a SyGuS solver.
The goal of \ourTool is to remove elements from the grammar and thus, by having a smaller search space, save time during synthesis.
To do this we combine two metrics, as shown in Figure~\ref{fig:model}: our predicted confidence that a grammar element is not needed, and our prediction of how much time will be saved by removing that element.
We focus on removing only elements where we are both confident that the grammar element is noncritical, and that removing the grammar element significantly impacts synthesis times.
By giving the constraints and the grammar definition to \ourTool, we predict which elements of the grammar can be safely removed.
By analyzing running times we predict which of these elements are benefical to remove.
We describe \ourTool in three sections, addressing dataset generation, the training stage, and our evaluation.

\section{Data Generation}

In order to learn a model for \ourTool, we need to generate a labelled dataset that maps constraints to grammar components in \grammarCrit.
This will allow us to predict, given a new set of constraints $\constraints '$, which grammar elements are noncritical for synthesis, and accordingly prune our grammar.
The generation of data for application to machine learning for program synthesis is a nontrivial problem, requiring careful construction of the dataset~\cite{shin2018synthetic}.
We break the generation of this dataset into two stages: first, we generate a set of programs, $\mathcal{P}$ from \grammar.
Then, for each program in $\mathcal{P}$, we generate constraints for that program.
We additionally need a dataset of synthesis times, in order to predict how long synthesis takes for a given set of constraints.

\subsection{Criticality Data}
To generate a set of programs $\mathcal{P}$, that can be generated from a grammar \grammar, we construct a synthesis query with no constraints.
We then run CVC4 with the command \texttt{--sygus-stream}, which instructs CVC4 to output as many solutions as it can find.
With no constraints, all functions satisfy the specification, and CVC4 will generate all permutations of (well-formed and well-typed) functions in the grammar, until the process is terminated (we terminate after generating $n$ programs).
Because CVC4 generates solutions of increasing size, we collect all generated programs, then shuffle the order to prevent data bias with respect to the order (size) in which CVC4 generated programs.

After generating programs, we generate corresponding constraints (in the form of input-output examples for PBE) for these functions.
To do this, for each program, $\program$, we randomly generate a set of inputs $I$, and compute the input-output pairs $\constraints = \{(i, \program(i))\ |\ i \in I\}$.
We then form a SyGuS problem $(\grammar, \constraints)$, where we know that the program $\program$ satisfies the constraints, and is part of the grammar: $\program \satisfies \constraints$ and $\program \in \grammar$.
This amounts to programs that \textit{could} be synthesized from the constraints (i.e. $(\grammar,\constraints) \synthesizes_{\infty} \program$).
It is important that our dataset represent programs that \textit{could} be synthesized, as opposed to what \textit{can} be synthesized (i.e. $(\grammar,\constraints) \synthesizes_{3600} \program$).
This is important because we will use this data set to try to learn the ``semantics'' of constraints, and we do not want to use this data set to additionally, inadvertently learn the limitations of the synthesis engine.

At this point, we have now constructed a dataset of triples of grammars (fixed for all benchmarks), constraints, and programs, $D = \{(\grammar, \constraints_1, \program_1) \ldots (\grammar, \constraints_n, \program_n)\}$.
In order to use $D$ to helps us predict \grammarCrit, we break up each triple by splitting each constraint set \constraints into its individual constraints.
For a triple $(\grammar,\constraints, \program)$, where $\constraints = \{c_1 \ldots c_m\}$, we generate a new set of triples $\{(\grammar, c_1, \program) \ldots (\grammar, c_m, \program)\}$.
The union of all these triples of individual constraints form our training set, \trainingCrit, that will be used to predict critical functions in the grammar for a given set of constraints.

\subsection{Timing Data}
In addition to a training set for predicting \grammarCrit, we also need a separate training set for predicting the time that can be saved by removing a terminal from the grammar.
This dataset maps grammar elements $\terminal \in \grammar$ to the effect on synthesis times, $\reals$, when $\terminal$ is dropped from the grammar.
To do this we require synthesis problems that more closely model the types of constraints that humans typically write.
We collect these set of benchmarks from users of the live coding interface for SyGuS~\cite{chi19}.
Because we had limited number of human-generated constraint examples, we augmented this with constraints generated from \trainingCrit.

We run synthesis for each problem with the full grammar, as well as with all grammars constructed by removing one element, $\terminal$.
For every synthesis problem benchmark, $1 \leq i \leq m$, we record the difference in synthesis times between running with the full grammar, and removing $\terminal$:

\begin{equation}
T^{\constraints_i}_\grammar - T^{\constraints_i}_{\drop{\grammar}{\terminal}}
\end{equation}

Thus, we create a training set, \trainingTime, relating each terminal $\terminal \in \setOf{\grammar}$ and a set of constraints, to the time it takes to synthesize a solution without that terminal.
\section{Training}

\subsection{Predicting criticality}

Our goal is to predict, given a set of constraints \constraints, if a terminal \terminal belongs to the set of terminals \setOf{\grammarCrit} for \constraints.
To do this, we use a Feedforward Neural Network (Multi-Layer Perceptron), with an extra embedding layer to encode the string valued input-output examples into feature vectors.
We train the neural network to predict the membership of each terminal $\terminal \in \setOf{\grammar}$ to the critical set $\setOf{\grammarCrit}$, based on a single constraint $c \in \constraints$.
This prediction produces a 1D binary vector of length $|\setOf{\grammar}|$, where 1 at position $i$ in the binary vector indicates the terminal in position $i$ is predicted to belong to the critical set. 

When a SyGuS problem has multiple ($|\constraints| \geq 2$) constraints, we run our prediction on each constraint individually.
We then use a voting mechanism to come to consensus on the construction of \grammarSynth.
After computing $|\constraints|$ binary vectors across all constraints, the vectors are summed to produce a final voting vector.
The magnitude of each element in this final voting vector represents the number of votes ``from each constraint'' that the terminal represented by that element is in the critical set.
We then use this final voting vector in combination with our time predictions.

\subsection{Predicting time savings}
It is only worthwhile to remove a terminal symbol \terminal from a grammar \grammar if
$T^{\constraints}_{\drop{\grammar}{\terminal}}$ is less than $T^{\constraints}_\grammar$.
If a \terminal stands to only give us a small gain in synthesis times, it may not be worth the risk that we incorrectly predicted its criticality.

To predict the amount of time saved by removing a terminal \terminal we examine the distribution of times in our training set \trainingTime.
For each terminal \terminal, we calculate $A_\terminal$, the average time increase that results from removing \terminal from the grammar.
Denoting the time to run $(\grammar, \constraints) \synthesizes \program$ as $T^\constraints_\grammar$, we can write $A_\terminal$ as:

\begin{equation*}
A_\terminal = \dfrac{\sum_{i = 1}^n  T^{\constraints_i}_\grammar - T^{\constraints_i}_{\drop{\grammar}{\terminal}}}{n}
\end{equation*}

If a terminal \terminal has a negative $A_\terminal$,
then removing it from the grammar actually slows down synthesis, on average.
As such, dropping the terminal from the grammar is not generally helpful.
Thus, we only consider those terminals with a positive $A_\terminal$ in our second step.

\subsection{Combining predictions}

With our predictions of the criticality a terminal \terminal and of time saved by removing \terminal, we must make a final decision on whether or not we should remove \terminal.
To do this, we take the top three terminals with the greatest average positive impact on synthesis time over the training set, as computed with $A_\terminal$. These tended to be terminals that mapped between types which saved more time due to the internal mechanisms and heuristics of the CVC4 solver.
We then use the final voting vector from our criticality prediction to choose only two out of the three to remove from \grammar to form \grammarSynth.
We chose to remove only two terminals from \grammar in order to minimize the likelihood of generating a \grammarSynth, such that $\setOf{\grammarSynth} \subseteq \setOf{\grammarCrit}$. We conjecture that the number of terminals removed is a grammar-dependent parameter that must be selected on a per grammar basis, just as the number of terminals with $A_\terminal > 0$ is grammar specific.

\subsection{Falling back to the full grammar}
\label{sec:fallingback}

There is some danger that \grammarSynth will, in fact, not be sufficient to synthesize a program.
Thus, we propose a strategy that
\begin{itemize}
	\item first, tries to synthesize a program with the grammar \grammarSynth
	\item second, if synthesis with \grammarSynth is unsuccessful, falls back to attempting synthesis with the full grammar \grammar.
\end{itemize} 

We determine how long to wait before switching from \grammarSynth to \grammar by finding an $x$ that minimizes:

\begin{equation}
\sum_{i = 1}^n \begin{Bmatrix}
 T^{C_i}_{\grammarSynth} & T^{C_i}_{\grammarSynth} < x\\ 
 \min(x+T^{C_i}_\grammar, t)& T^{C_i}_{\grammarSynth} >  x
\end{Bmatrix}
\end{equation}
where $C_1 \ldots C_n$ are the constraints from the training set, and $t$ is the timeout for synthesis. 

Ideally, as captured in the first line of the sum, $(\constraints_i, \grammarSynth) \synthesizesInTime{x} P$ will finish before  $T^{C_i}_{\grammarSynth}$ = x.
However, if a benchmark does not finish in that time, it will fall back on the full grammar.
Then, either $(\constraints_i, \grammarSynth) \synthesizesInTime{t - x} P$ will succeed, and  synthesize the expression in total time $x+T^{C_i}_{\grammar}$, or synthesis will timeout, in total time $(t - x) + x = t$.

\section{Experiments}

\begin{figure*}[t!]
   \begin{tikzpicture}
      \begin{axis}[
      width  = 0.99*\textwidth,
      height = 7cm,
      major x tick style = transparent,
      ybar=2*\pgflinewidth,
      bar width=2.5pt,
      ymajorgrids = true,
      symbolic x coords={
                phone-8.sl,
                phone-6.sl,
                phone-5.sl,
                phone-9_short.sl,
                phone-10_short.sl,
                phone-9.sl,
                phone-10.sl,
                lastname-long.sl,
                lastname-long-repeat.sl,
                phone-6-long-repeat.sl,
                phone-5-long-repeat.sl,
                phone-7-long.sl,
                phone-7-long-repeat.sl,
                phone-5-long.sl,
                phone-8-long-repeat.sl,
                phone-9-long-repeat.sl,
                phone-6-long.sl,
                phone-8-long.sl,
                phone-10-long-repeat.sl,
                phone-10-long.sl
      },
      xtick = data,
      xticklabels={ 43,44,45,46,67,48,49,50,51,52,53,54,55,56,57,58,59,60,61,62
      },
      xticklabel style={align=center},
      scaled y ticks = false,
      ymin=0,
      ylabel=Seconds to complete,
      xlabel=Benchmark Id,
      legend style={at={(0.1,0.99)},anchor=north},
  ]

      \addplot[style={fill=white},error bars/.cd, y dir=both, y explicit]
          coordinates {
                        (phone-8.sl, 4.720944166)
                        (phone-6.sl, 4.851982832)
                        (phone-5.sl, 4.8752141)
                        (phone-9_short.sl, 4.882001162)
                        (phone-10_short.sl, 8.80894804)
                        (phone-9.sl, 12.081002)
                        (phone-10.sl, 31.22909999)
                        (lastname-long.sl, 32.39816236)
                        (lastname-long-repeat.sl, 32.48848486)
                        (phone-6-long-repeat.sl, 83.59451318)
                        (phone-5-long-repeat.sl, 84.76578879)
                        (phone-7-long.sl, 87.83005524)
                        (phone-7-long-repeat.sl, 89.1325171)
                        (phone-5-long.sl, 90.80672789)
                        (phone-8-long-repeat.sl, 91.03858423)
                        (phone-9-long-repeat.sl, 91.18507075)
                        (phone-6-long.sl, 98.14970112)
                        (phone-8-long.sl, 108.064276)
                        (phone-10-long-repeat.sl, 149.532891)
                        (phone-10-long.sl, 153.3208489)
                  };

      \addplot[style={fill=black},error bars/.cd, y dir=both, y explicit,error bar style=red]
           coordinates {
                            (phone-8.sl, 2.165008783)
                            (phone-6.sl, 1.967978001)
                            (phone-5.sl, 2.202147007)
                            (phone-9_short.sl, 4.725084066)
                            (phone-10_short.sl, 8.279015064)
                            (phone-9.sl, 4.862329006)
                            (phone-10.sl, 8.491020918)
                            (lastname-long.sl, 25.48625207)
                            (lastname-long-repeat.sl, 24.91945791)
                            (phone-6-long-repeat.sl, 25.30598307)
                            (phone-5-long-repeat.sl, 33.68135405)
                            (phone-7-long.sl, 26.15487099)
                            (phone-7-long-repeat.sl, 26.22752309)
                            (phone-5-long.sl, 30.00700617)
                            (phone-8-long-repeat.sl, 35.64468002)
                            (phone-9-long-repeat.sl, 77.01791883)
                            (phone-6-long.sl, 24.75411797)
                            (phone-8-long.sl, 29.94093895)
                            (phone-10-long-repeat.sl, 129.425879)
                            (phone-10-long.sl, 133.2229772)
           };

  \legend{CVC4, \ourToolCVC}
  \end{axis}
  \end{tikzpicture}
  \caption{The top 20 problems with longest synthesis time for CVC4 (excepting timeouts), and the corresponding synthesis times for \ourToolCVC.}
  \label{fig:top20}
\end{figure*}
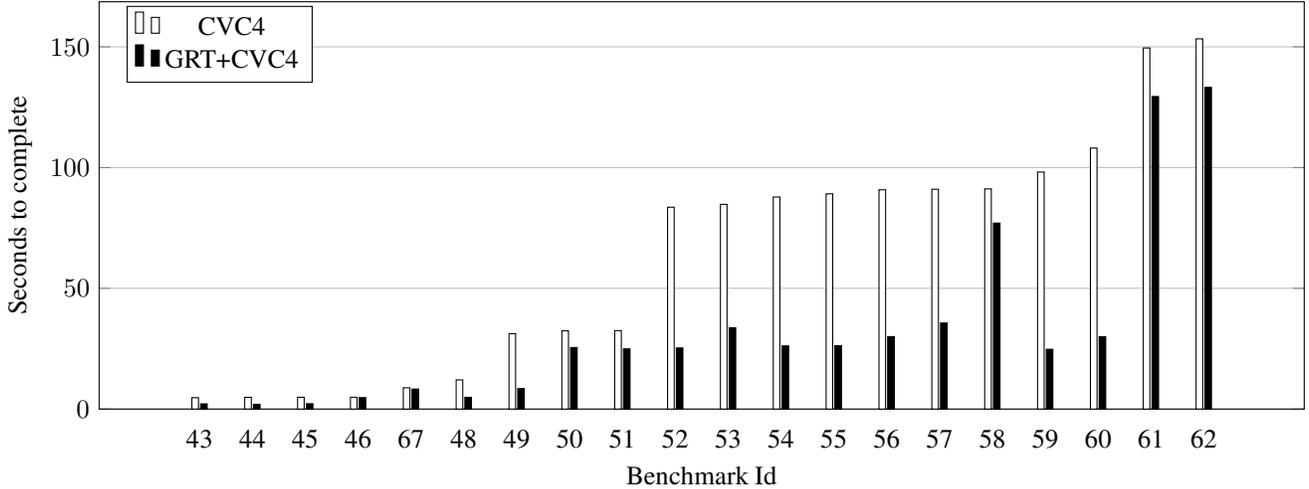

The SyGuS competition~\cite{sygusCompetition} provides public competition benchmarks and results from previous years.
In particular, the PBE Strings dataset provides a collection of PBE problems over a grammar that includes string, integer, and Boolean manipulating functions.
First, we describe our approach to generating a training set of PBE problems over strings.
Then, we present our results running \ourTool against the 2019 competition's winner in the PBE Strings track, CVC4~\cite{notzli2019syntax,barrett2011cvc4,sygusCompetition2019}.
We are able to reduce synthesis time by $47.65\%$ and synthesize a new solution to a benchmark that was left unsolved by CVC4.

\subsection{Technical details}

The data triples generated during our initial data generation process of \trainingCrit \, are triples of strings.
However, the neural network cannot process input-output pairs of type string as input. 
Thus, this data must be encoded numerically before it can be utilized to train the neural network. 
Each character in the input-output pairs is converted to its ASCII equivalent integer value. 
The size of each pair is then standardized by adding a padding of zeros to the end of each newly encoded input and output vector respectively. 
This creates two vectors: the encoded input and the encoded output, both of which have a length of 20.
These two vectors are then concatenated to give us a single vector for training. 
By the end of this process the triples created in our first data generation step are now one vector of type $\mathbb{N}^{40}$ representing the input-output pair and a correct label $P$ that will be predicted.

To generate the training set for predicting synthesis times, \trainingTime, we combine human generated and automatically generated SyGuS problems.
Specifically, we use 10 human generated SyGuS problems, and 20 randomly selected problems from \trainingCrit.

The overall architecture of our model can be categorized as a multi-layer perceptron (MLP) neural network.
More specifically, our model is made up of five fully connected layers: the input layer, three hidden layers, and the output layer.
By using the Keras Framework, we include an embedding layer along with our input layer which enables us to create unique vector embeddings of length 100 for any given input-output pair in the dataset.
This embedding layer learns the optimal weights used to create these unique vectors through the training process. 
Thus, we create an encoding of the input-output pairs for training, while simultaneously standardizing the scale of the vector before it reaches the first hidden layer.
The hidden layers of the model are all fully connected, and all use the sigmoid activation function. 
In addition, we implement dropout during training to ensure that overfitting does not occur.
The size of the hidden layers was calculated using a geometric series to ensure that there was a consistent decrease in layer size as the layers get closer to the output layer.
Specifically, the size of each hidden layer was calculated by:

\begin{equation}
\mathit{HL}_{\mathit{size}}(n) = {\mathit{input}}_{\mathit{size}} \big( \frac{\mathit{output}_{\mathit{size}}}{\mathit{input}_{\mathit{size}}} \big) ^ {\frac{n}{L_{\mathit{num}}+1}}
\end{equation}

where $L_{num}$ represents the total number of layers in the network.
Our model used the Adam optimization method and the binary-cross entropy loss function as it is well suited for multi-label classification. Overall, our model was trained on 124928 data points for 15 epochs with a batch size of 200 producing a training time of 228 seconds.

\subsection{Results}
After generating our data sets and training our model, we wrote a wrapper script to run \ourTool as a preprocessor for CVC4's SyGuS engine.
We compared the synthesis results of \ourToolCVC  with the synthesis results of running CVC4 alone.
All experiments were run on MacBook Pro with a 2.9 GhZ Intel i5 processor with 8GB of RAM.
CVC4 uses a default random seed, and is deterministic over the choice of that seed, so the results of synthesis from CVC4 on a given grammar and set of constraints are deterministic.
We note that our training data in no way used any of the SyGuS benchmarks.

\ourToolCVC outperformed directly calling CVC4 on 32 out of 64 benchmarks (50\%), with a reduction in total synthesis time over all benchmarks from 1304.87 seconds with CVC4 to 683.09 seconds with \ourToolCVC.
On one benchmark, CVC4 timed out and was not able to find a solution (even when the timeout was increased to 5000 seconds), while \ourToolCVC found a solution within the timeout specified by the SyGuS Competition rules (3600 seconds).
On one benchmark, both CVC4 and \ourToolCVC timeout (TO) and are not able to find a solution.
On the other 31 benchmarks, CVC4 performed the same (within $\pm0.1\text{s}$) with and without the preprocessor.
All the benchmarks for which CVC4 performed the same as \ourToolCVC finish in under 2 seconds, and 28 of the 31 finish in under a second.
In these cases there was little room for improvement even with \ourToolCVC.

Figure~\ref{fig:top30} shows the exact running times with both the full and reduced grammars from the benchmarks with the 30 largest running times with the full grammar.
These are the benchmarks for which the synthesis times and size of the solution diverge most meaningfully, however all other data is available in the supplementary material for this paper.
Figure~\ref{fig:top30} also shows $|P|$ and $|P^*|$, the sizes of the programs found by the CVC4 and \ourToolCVC, respectively.
We define size of a program as the number of nodes in the abstract syntax tree of the program.
In terms of the grammar \grammar, this is the number of terminals (including duplicates) that were composed to create the program.

\begin{figure}
   \begin{tikzpicture}
      \begin{axis}[
      width  = 0.49*\textwidth,
      height = 6cm,
      major x tick style = transparent,
      ybar=2*\pgflinewidth,
      bar width=2.5pt,
      ymajorgrids = false,
      symbolic x coords={
        phone-10.sl,
        phone-10-long-repeat.sl,
        phone-10-long.sl,
        phone-9-long-repeat.sl,
        phone-9.sl
      },
      xtick = data,
      xticklabels={ 49, 61, 62, 58, 48 
      },
      scaled y ticks = false,
      enlarge x limits=0.30,
      ymin=0,
      ylabel=Length of sythesized solutions,
      xlabel=Benchmark Id,
      legend style={at={(0.75,0.99)},anchor=north},
  ]

      \addplot[style={fill=white},error bars/.cd, y dir=both, y explicit]
          coordinates {
              (phone-10.sl, 97)
              (phone-10-long-repeat.sl, 49)
              (phone-10-long.sl, 49)
              (phone-9-long-repeat.sl, 47)
              (phone-9.sl, 56)
          };

      \addplot[style={fill=black},error bars/.cd, y dir=both, y explicit,error bar style=red]
           coordinates {
              (phone-10.sl, 49)
              (phone-10-long-repeat.sl, 65)
              (phone-10-long.sl, 65)
              (phone-9-long-repeat.sl, 50)
              (phone-9.sl, 52)
           };

  \legend{CVC4, \ourToolCVC}
  \end{axis}
  \end{tikzpicture}
  \caption{When the \ourToolCVC found a different solution than CVC4, it was on average shorter than the solution found with the full grammar.}
  \label{fig:sizes}
\end{figure}
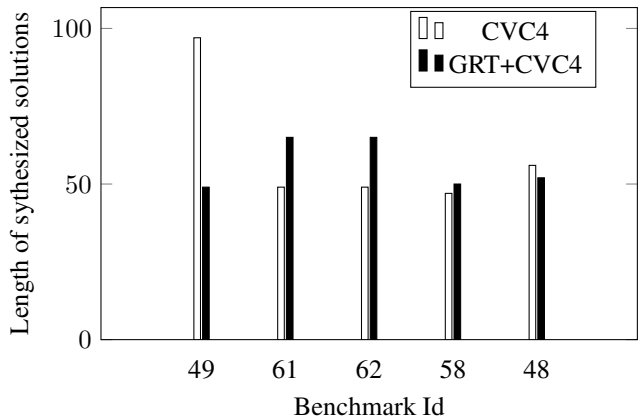

\begin{figure}[t]
    \footnotesize
    \def\arraystretch{1.1}
    \setlength{\tabcolsep}{0.55em}
    \begin{tabular}{c|c|c|c|c|c}%
    \bfseries id & \bfseries file & $T^{\constraints}_\grammar$ & $T^{\constraints}_{\grammarSynth}$ & $|P|$ & $|P^\star|$
    \csvreader[no head]{localResults.csv}{}
    {\\\hline \csvcoli&\csvcolii&\csvcoliii&\csvcoliv&\csvcolv&\csvcolvi}
    \\\hline
    \end{tabular}
    \caption{Synthesis results over the 30 longest running benchmarks from SyGuS Competition's PBE Strings track.}
    \label{fig:top30}
\end{figure}

In Figure~\ref{fig:top20}, we present a visual comparison of the results for the 20 functions that took CVC4 the longest, while still finishing in the 3,600 second time limit.
We note that we have the largest gains on the problems for which CVC4 is the slowest.
Problems that CVC4 already handles quickly stand to benefit less from our approach.

In order to get a better baseline to understand the impact of GRT on running times, we ran a version of GRT with only the criticality prediction, which we call GRTC.
In this case, GRTC+CVC4 actually performed worse than CVC4 by itself, increasing the running time on 53 out of the 62 benchmarks that did not timeout on CVC4.

On all but 5 benchmarks, CVC4 synthesized the same program when running with \grammar and \grammarSynth.  
The sizes of the programs (in terms of the number of terminal symbols used) for the benchmarks on which CVC4 synthesized different programs are shown in Figure~\ref{fig:sizes}.
While on some benchmarks \ourToolCVC produced a larger solution than CVC4, as a whole the sum of the size of all solutions for CVC4 was 806, while for \ourToolCVC it was 789.
Thus, overall, we were able to outperform CVC4 on size of synthesis as well.

The SyGuS competition scores each tool using the formula: $5N+3F+S$, where $N$ is the number of benchmarks solved (non-timeouts), $F$ is based on a ``pseudo-logarithmic scale''~\cite{sygusCompetition} indicating speed of synthesis, and $S$ is based on a ``pseudo-logarithmic scale'' indicating size of the synthesized solution.
On all three of these measurements, \ourToolCVC performed better than CVC4.
There are number of other synthesis tracks available in the SyGuS competition, which do not involve PBE constraints. 
We note that our approach can selectively be applied as a preprocessing step for input in the PBE track without incurring an overhead on other synthesis tasks.

Although we implemented a strategy to manage a switch from the reduced grammar back to the full grammar, we found in practice that the optimal strategy for our system was to exclusively use the reduced grammar.
Because we had conservatively pruned the grammar, we had no need to switch back to the full grammar. 

\section{Conclusions}

In a way, by training on a dataset we generate from the output of the interpreter of the language, we are encoding an approximation of the semantics into our neural network.
While the semantic approximation is too coarse to drive synthesis itself, we can use it to prune the search space of potential programs.
By predicting terminals impact on synthesis time, we more conservatively remove only terminals likely to have a positive impact.
In conjunction with analytically driven tools, we can then significantly improve synthesis times with very little overhead.

While we have presented \ourTool, which demonstrates a significant gain in performance over all existing SyGuS solvers, we still have many opportunities for further improvement.
In our prediction of the potential time saved by removing a terminal from the grammar, we have simply used the average expected value over all samples in the dataset.
By using a neural network here, we may be able to leverage some property of the SyGuS problem constraints to have more accurate potential time savings predictions.
This would allow us, possibly in combination with a more advance prediction combination strategy, to more aggressively prune the grammar.
The drawback to this approach is that we may then potentially remove too much from the grammar.
One of the key features of \ourTool is that it introduces no new timeouts, that is, it does not remove any critical parts of the grammar.

Additionally, our prediction of criticality of a terminal uses a voting mechanism to combine the prediction based on each constraint.
While this worked well in practice, this strategy ignores the potential for interaction between constraints.
In our preliminary exploration, we were not able to construct a model that captures this inter-constraint interaction in a useful way.
This may be a path for future work.
In a similar vein, there exist a number of other works that define a criticality measure for each terminal in the SyGuS grammar~\cite{BalogGBNT17,kalyan2018neural}.
It may be possible to leverage these in place of our criticality measure, and in combination with our time savings prediction, to achieve better results.

So far we have only explored the PBE Strings track of the SyGuS Competition.
The competition also features a PBE BitVectors track where our technique may have significant gains as well.
This would require a new encoding scheme, but the overall approach would remain similar.
In general, extending this work to allow for other PBE types, as well as more general constraints, would broaden the potential real-world application of SyGuS.

\paragraph{Acknowledgments}
This work was supported in part by NSF grants CCF-1302327, CCF-1715387, and CCF-1553168.

\bibliographystyle{aaai} 
\bibliography{bibfile}

\end{document}